\pdfoutput=1

\documentclass[11pt]{article}

\usepackage[final]{EACL2023}

\definecolor{c1}{HTML}{ed1941}
\definecolor{c2}{HTML}{2485a6}

\usepackage{times}
\usepackage{latexsym}
\usepackage{threeparttable}
\usepackage{multirow}
\usepackage{booktabs} 
\usepackage{amssymb}
\usepackage{amsmath}
\usepackage{tabu}
\usepackage{hyperref}
\usepackage{colortbl}
\usepackage{graphicx}
\usepackage{bbm}
\usepackage{makecell}
\usepackage[T1]{fontenc}
\newcommand{\ignore}[1]{}
\newcommand{\fms}[1]{}
\newcommand{\lihu}[1]{}
\usepackage[textsize=tiny, disable]{todonotes}
\def\gv#1{\todo[color=green]{#1}}

\usepackage[T1]{fontenc}

\usepackage[utf8]{inputenc}

\usepackage{microtype}

\usepackage{inconsolata}

\title{GLADIS: A General and Large Acronym Disambiguation Benchmark}

\author{
Lihu Chen\textsuperscript{\rm 1},
Gaël Varoquaux\textsuperscript{\rm 2}, 
Fabian M. Suchanek\textsuperscript{\rm 1} \\
\textsuperscript{\rm 1} LTCI \& Télécom Paris \& Institut Polytechnique de Paris, France \\
\textsuperscript{\rm 2} Soda, Inria Saclay, Université Paris-Saclay, France \\
\texttt{\{lihu.chen, fabian.suchanek\}@telecom-paris.fr}\\ 
\texttt{\{gael.varoquaux\}@inria.fr}
}

\begin{document}
\maketitle
\begin{abstract}
Acronym Disambiguation (AD) is crucial for natural language understanding on various sources, including biomedical reports,  scientific papers, and search engine queries. 
However, existing acronym disambiguation
benchmarks and tools are limited to specific domains, and the size of prior benchmarks is rather small. 
To accelerate the research on acronym disambiguation, we construct a new benchmark named GLADIS with three components: (1) a much larger acronym dictionary with 1.5M acronyms and 6.4M long forms; (2) a pre-training corpus with 160 million sentences;
(3) three datasets that cover the
general, scientific, and biomedical domains.
We then pre-train a language model, \emph{AcroBERT}, on our constructed corpus for general acronym disambiguation, 
and show the challenges and values of our new benchmark.
\end{abstract}

\section{Introduction}

An acronym is an abbreviation formed from the initial letters of a longer name. 
For instance, the following two sentences contain the acronym \textit{``AI''}: 
(1) \textit{This is the product's first true \textbf{AI} version, and it understands your voice instantly.} 
(2) \textit{In the United States, the \textbf{AI} for potassium for adults is 4.7 grams.}
The long forms (or expanded forms) for the same acronym are \textit{``Artificial Intelligence''} and \textit{``Adequate Intake''}, respectively.

\begin{table}[t] 
	\centering 
	\scriptsize
	\setlength{\tabcolsep}{2.2mm}{
		\begin{threeparttable} 
			\begin{tabular}{cccc}  
				\toprule  
				\textbf{ID}&\textbf{Long Form}&\textbf{Popularity}&\textbf{Domain}\cr
				\midrule
				1&{\textbf{\textit{Artificial Intelligence}}}&$\bigstar\bigstar\bigstar\bigstar\bigstar$&Computer Science\cr
				2&\textcolor{c1}{\textit{Adequate Intake}}&$\bigstar\bigstar\bigstar\bigstar$&Food and Nutrition\cr
				3&\textcolor{c1}{\textit{Aromatase Inhibitor}}&$\bigstar\bigstar\bigstar$&Chemistry\cr
				4&\textcolor{c1}{\textit{Apoptotic Index}}&$\bigstar\bigstar\bigstar$&Biomedicine\cr
				5&\textcolor{c1}{\textit{Asynchronous Irregular}}&$\bigstar\bigstar\bigstar$&Neuroscience\cr
				6&\textcolor{c1}{\textit{Amnesty International}}&$\bigstar\bigstar$&Organization\cr
				7&\textcolor{c1}{\textit{Anterior Insula}}&$\bigstar\bigstar$&Biomedicine\cr
				8&\textcolor{c1}{\textit{Air India}}&$\bigstar\bigstar$&Organization \cr
				9&\textbf{\textit{Article Influence}}&$\bigstar\bigstar$&Science \cr
				......&\cr
				2243&\textcolor{c1}{\textit{Agricultural Implement}}&$\bigstar$&Agriculture \cr
				\bottomrule
			\end{tabular}
			\caption{Long form candidates for the acronym \textit{``AI''} from our acronym dictionary. The SciAD benchmark \cite{veyseh2020does} only includes two long terms (black) in the scientific domain.  The popularity is the occurrence frequency in our collected corpora.} 	\label{tab:example_of_ai}
		\end{threeparttable}
	}
\end{table}
Acronym Disambiguation (AD) is the task of mapping a given acronym in a given sentence to the intended long form. Acronym disambiguation is crucial for downstream tasks such as information extraction, machine translation, and query analysis in search engines~\cite{jain2007acronym,islamaj2009understanding}. Acronym disambiguation is also important for humans: acronyms may make a text more difficult to understand for readers who are not familiar with the specific domain.
A study on a Microsoft question answering forum found that only 7\% of the acronyms co-occur with their corresponding long forms, which confuses the readers about the meaning of a text \cite{li2018guess}. 

Acronym Disambiguation has received more attention in the past few years.
The first step in acronym disambiguation is usually the creation of a dictionary, i.e., a mapping of each acronym to one or more long forms. Early systems extracted acronyms and their definitions automatically from texts by rule-based \cite{schwartz2002simple} or supervised \cite{nadeau2005supervised} methods.
Once a dictionary is available, acronym disambiguation methods expand acronyms in a given text by capturing the contexts for specific domains, e.g., the enterprise domain \cite{li2018guess}, biomedical texts \cite{jin2019deep}, and scientific papers \cite{charbonnier2018using}. 
Madog \cite{veyseh2021maddog} was the first general and web-based system,
recognizing and expanding acronyms across multiple domains.
Several benchmarks have also been constructed, including for the
biomedical area \cite{suominen2013overview} and the
scientific area \cite[SciAD,][]{veyseh2020does}.
Several methods fine-tuned SciBERT \cite{beltagy2019scibert} on SciAD to disambiguate acronyms in scientific documents \cite{pan2021bert, zhong2021leveraging, li2021simclad}. 

Although these works have significantly advanced the progress of acronym disambiguation, they suffer from three main limitations.
First, most existing dictionaries (and benchmarks) focus on one specific domain. In real-world applications, however, the input text may be general, cross-domain, or of an unspecified domain (as in search engine queries). Second, existing dictionaries are limited in size. 
For example, there are only two long forms for the acronym \textit{``AI''} in SciAD (Table~\ref{tab:example_of_ai}), which is constructed from arXiv.
However, we find that the two long forms \textit{``Asynchronous Irregular''} and \textit{``Anterior Insula''} also appear in scientific papers on arXiv \cite{girardi2019self, vadovivcova2014affective}, and the acronym \textit{``AI''} also appears separately without the long form in sentences.
In our work, we actually find 
at least 2\,243 different long forms for \textit{``AI''}. 
Besides, SciAD suffers from the problem of data leakage, because the train and test sets have overlapping pairs of acronym and long form.
Finally, current general AD systems such as MadDog \cite{veyseh2021maddog} rely on static word embeddings and LSTMs (Long Short Term Memory \cite{hochreiter1997long}). Thus, they do not leverage pre-training on large corpora, which drives the current state of the art in most NLP tasks with contextual embeddings like BERT \cite{devlin2018bert}.

With this work, we aim to improve Acronym Disambiguation along two dimensions:
First, we automatically construct GLADIS, a \textbf{G}eneral and \textbf{L}arge \textbf{A}cronym \textbf{DIS}ambiguation benchmark that includes a larger dictionary, a pre-training corpus and three datasets covering the general, biomedical, and scientific domains.
Our dictionary contains 1.5M acronyms and 6.4M long forms, which trumps existing dictionaries by a factor of 3. 
We complement this dictionary by three domain-specific datasets for acronym disambiguation, which are adapted from three existing human-annotated and crowd-sourced datasets~\cite{mohan2018medmentions,onoe2020fine,veyseh2020does}.
The pre-training corpus has 160 million sentences with acronyms, collected from the Pile dataset~\cite{gao2020pile} with a rule-based algorithm~\cite{schwartz2002simple}.
Second, we propose AcroBERT, the first pre-trained language model for general acronym disambiguation. Our experiments show that this model 
outperforms existing systems across multiple domains. 
Our code and data are available at \url{https://github.com/tigerchen52/GLADIS}.

\ignore{
The rest of this paper is structured as follows: 
Section~\ref{sec:related_work} discusses related
work, Section~\ref{sec:dictionary_construction} introduces the construction of our acronym dictionary, Section~\ref{sec:benchmark} presents our new acronym disambiguation benchmark, Section~\ref{sec:acrobert} discusses our new model (AcroBERT), Section~\ref{sec:experiment} shows the experimental results, and Section~\ref{sec:conclusion} concludes. Section~\ref{sec:limitations} discusses limitations of our approach.

Appendix~\ref{sec:details} contains the details of the experimental settings, and Appendix~\ref{sec:additional} contains additional experiments. All code and data will be made publicly available.
}

\section{Related Work}\label{sec:related_work}

\subsection{Acronym Identification and Disambiguation}
To expand acronyms, there are usually two sub-tasks: Acronym Identification (AI), which creates a dictionary of acronyms and their definitions from a given document, and Acronym Disambiguation (AD), which aims to link acronyms in the input text to the correct long forms from a dictionary.

The study of acronym identification has a long history. Early work observed that acronyms and their long forms appear frequently together in a document, as in \textit{``Artificial Intelligence (AI)''}.
Based on this pattern, many approaches identify and extract acronyms by using rules \cite{yeates2000using,larkey2000acrophile, pustejovsky2001automatic, park2001hybrid, yu2002mapping, schwartz2002simple, adar2004sarad,ao2005alice,okazaki2006building,sohn2008abbreviation, veyseh2021maddog} or supervised methods \cite{chang2002creating, nadeau2005supervised, kuo2009bioadi, movshovitz2012alignment, liu2017multi, wu2017long, zhu2021bert}. In our work, we build on previous work \cite{schwartz2002simple} for Acronym Identification, and focus mainly on disambiguation.

As for acronym disambiguation, early solutions manually designed features to score each pair of acronyms and long forms, by either unsupervised \cite{jain2007acronym, henriksson2014synonym} or supervised machine learning \cite{pakhomov2005abbreviation, yu2007using,stevenson2009disambiguation,finley2016towards, li2018guess}. 
Later, deep learning approaches were introduced to the task, using embeddings to represent word sequences. 
The methods can be categorized as static embedding-based \cite{wu2015clinical,li2015acronym, charbonnier2018using} and dynamic embedding-based \cite{jin2019deep,pan2021bert, zhong2021leveraging, li2021simclad}, where the former generates fixed representations for words in a pre-defined vocabulary and the latter can represent arbitrary words dynamically based on specific contexts. 
One main limitation of these methods is that they are domain-specific systems that can be applied only to a certain field such as the biomedical domain or scientific documents.
To generalize the system, \citet{ciosici2018abbreviation} presented the Abbreviation Expander,  \citet{veyseh2021maddog} proposed MadDog, and~\citet{pereira2022acx} developed AcX, all of which can be used in multiple domains. 
In this paper, we improve over the performance of these systems by adapting transformer-based methods and pre-training strategies.

\begin{figure*}[t]
	\centering
	\includegraphics[width=1.0\textwidth]{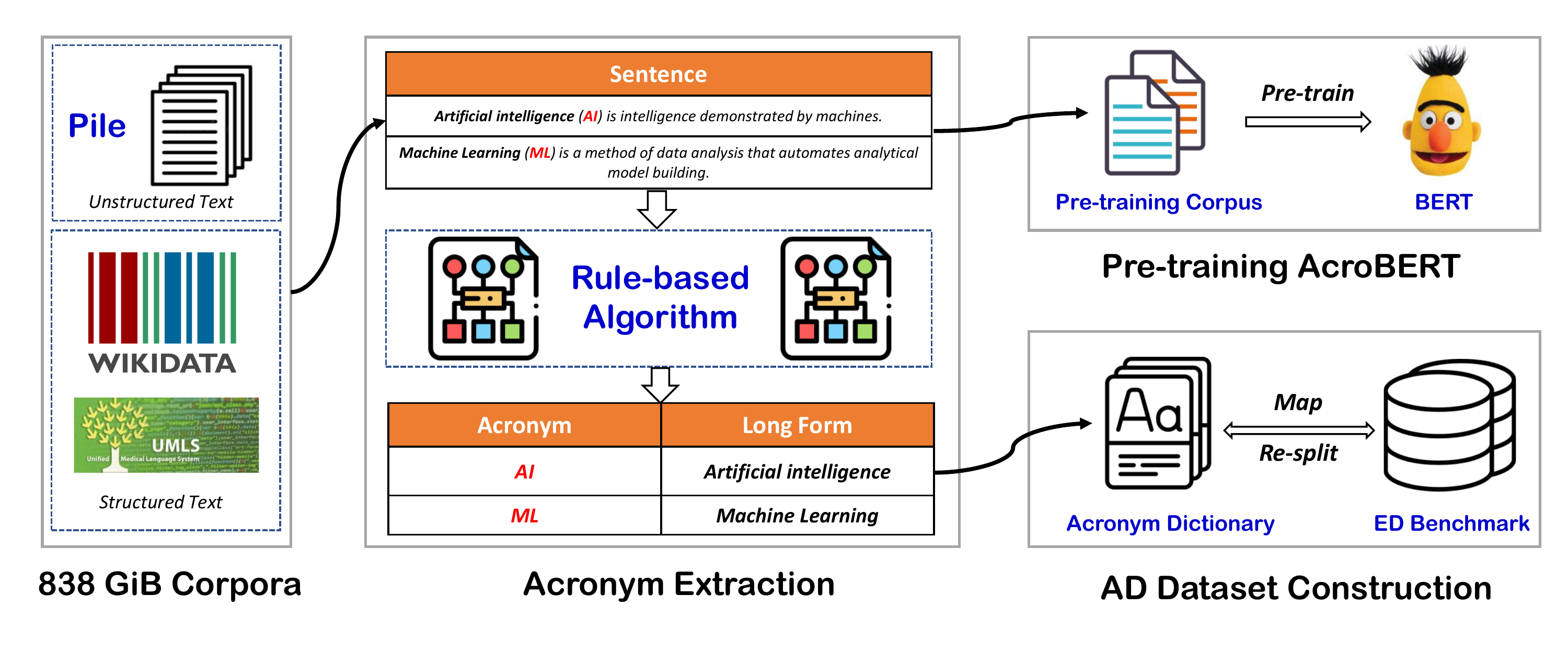}
	\caption{Framework of our benchmark construction. The \textit{``ED''} in the lower right corner means \textit{``Entity Disambiguation''}.}
	\label{fig:benchmark_constrcution}
\end{figure*}

\subsection{Existing benchmarks}\label{sec:relwork-bm}

Most current public datasets for acronym expansion are focused on a particular domain,
such as the biomedical domain~\cite{suominen2013overview,wen2020medal} or science~\cite{charbonnier2018using, veyseh2020does}.
Some works adopt two domain-specific datasets for better evaluations \cite{ciosici2019unsupervised, veyseh2022macronym}.
The main limitation of these benchmarks is two-fold: first,
their acronym dictionaries are rather small.
For instance, the average number of candidates per acronym in the SciAD benchmark \cite{veyseh2020does} is 3.15 while in our benchmark the number is greater than 200.
Second,
there are no AD evaluation sets that cover multiple domains.
We also note that, in SciAD, the train and test sets have overlapping pairs of acronym and long form. For example, the pair \emph{$\langle$CT, Computed Tomography$\rangle$} appears in the training, validation, and test sets.

\ignore{
	\begin{figure*}[t]
		\centering
		\includegraphics[width=1.0\textwidth]{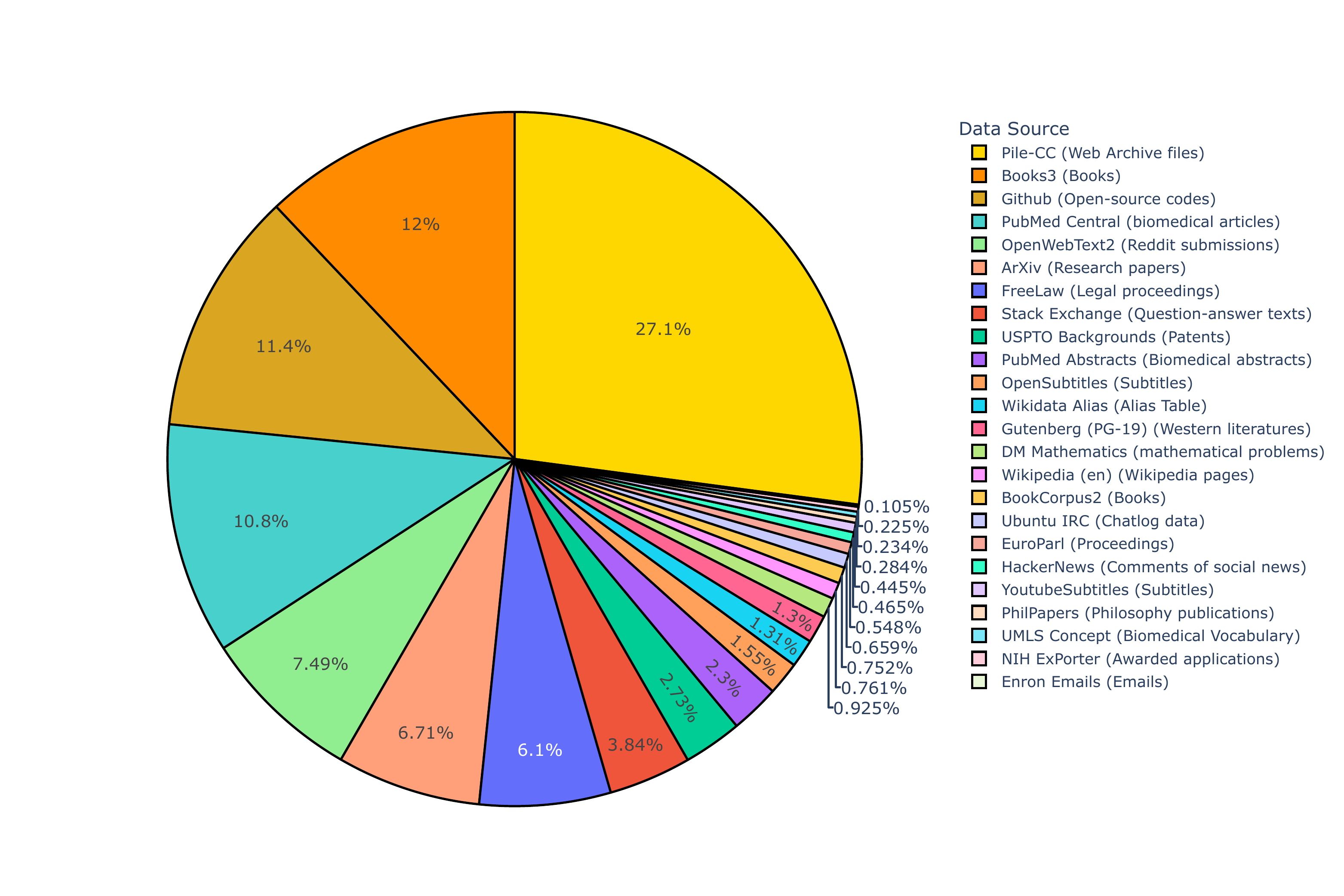}
		\caption{Corpora sources for acronym extraction (838 GiB in total). Apart from Wikidata Alias \cite{vrandevcic2014wikidata} and UMLS Concept \cite{bodenreider2004unified}, other corporas are from Pile \cite{gao2020pile}. The legend shows the name and domain of each corpora.}
		\label{fig:data_source}
	\end{figure*}
}

\section{Constructing GLADIS} \label{sec:dictionary_construction}%
Our GLADIS benchmark consists of three components: a dictionary, a pre-training corpus, and three domain-specific datasets.

\subsection{Dictionary and Pre-training Corpus}

We propose an acronym dictionary that addresses the shortcomings of existing dictionaries (Section~\ref{sec:relwork-bm}) by being (1) cross-domain and (2) large in size. To construct this dictionary, we apply rule-based extraction on a large set of corpora that contain acronym definitions. In this process, we can also obtain a large number of sentences containing acronyms as the pre-training corpus.

\begin{table}[ht] 
	\centering 
	\scriptsize
	\setlength{\tabcolsep}{2.2mm}{
		\begin{threeparttable} 
			\begin{tabular}{cccc}  
				\toprule  
				Subset&Domain&Size (GiB)\cr
				\midrule
				Pile-CC  &Web Archive files &227.12\cr
				Books3  &Books&100.96\cr
				Github  &Open-source codes&95.16\cr
				PubMed Central  &Biomedical articles&90.27\cr
				OpenWebText2  &Reddit submissions&62.77\cr
				ArXiv  &Research papers&56.21\cr
				FreeLaw  &Legal proceedings&51.15\cr
				Stack Exchange  &Question-answer texts&32.20\cr
				USPTO Backgrounds  &Patents&22.90\cr
				PubMed Abstracts  &Biomedical abstracts&19.26\cr
				OpenSubtitles  & Subtitles&12.98\cr
				Gutenberg (PG-19)  &Western literatures&10.88\cr
				DM Mathematics  &mathematical problems&7.75\cr
				Wikipedia (en)  &Wikipedia pages&6.38\cr
				BookCorpus2  &Books&6.30\cr
				Ubuntu IRC  &Chatlog data&5.52\cr
				EuroParl  &Proceedings&4.59\cr
				HackerNews  &Comments of social news &3.90\cr
				YoutubeSubtitles  &YouTube subtitles&3.73\cr
				PhilPapers  &Philosophy publications&2.38\cr
				NIH ExPorter  &Awarded applications&1.89\cr
				Enron Emails  &Emails&0.88\cr
				Wikidata Alias &Alias Table&11.00\cr
				UMLS Concept &Biomedical Vocabulary&1.96\cr
				Total&-&838.14\cr
				\bottomrule
			\end{tabular}
			\caption{Sources for acronym extraction. All corpora except Wikidata Alias and UMLS Concept are from Pile \cite{gao2020pile}. } 	\label{tab:data_source}
		\end{threeparttable}
	}
\end{table}

\begin{table*}[ht] 
	\centering 
	\small
	\setlength{\tabcolsep}{2.2mm}{
		\begin{threeparttable} 
			\begin{tabular}{ccccccc}  
				\toprule  
				&train&valid&test&unique short form&long forms per acronym&overshadowed ratio  \cr
				\midrule
				General&13,269&7,024&7,125&1,147&248&29.8\%\cr
				Scientific&28,023&14,134&14,066&2,922&262&68.7\%\cr
				Biomedical&6,295&3,150&3,149&2,909&278&27.4\%\cr
				\bottomrule
			\end{tabular}
			\caption{Statistics of our new Acronym Disambiguation Benchmark. The last column shows the ratio of overshadowed samples in the dataset: long forms with the same acronym but not the most popular one.} 	\label{tab:stat_benchmark}
		\end{threeparttable}
	}
\end{table*}

\textbf{Input Corpora.} For the textual data source, we use the Pile dataset \cite{gao2020pile}, an 825 GiB English corpus constructed from 22 diverse high-quality subsets (see the details of Pile in Appendix~\ref{sec:pile}). We also make use of structured knowledge from knowledge bases, namely 
the Alias Table from Wikidata and the Concept Names from UMLS. Both of them contain alternate names for canonical entities, and these may be acronyms or not.
To consider only the acronyms, we produce pairs of the canonical name and an alternate name in the form \textit{``canonical form (alternate name)''}. The rule-based algorithm will then decide whether to extract an acronym or not.
Table~\ref{tab:data_source} shows the statistics of our sources. They cover a wide range of domains including Web pages, books, scientific and biomedical papers, legal documents, etc.  

\textbf{Acronym Extraction.} To extract acronyms from the textual sources, we use the rule-based algorithm proposed by \citet{schwartz2002simple}. It assumes that acronyms follow a predictable pattern, e.g., \textit{long form ( acronym )} or  \textit{acronym ( long form )},
and then uses rules to extract candidate pairs by identifying parentheses and surrounding tokens.
Experimental results show that this simple algorithm achieves 95\% precision and 82\% recall, averaged over two datasets.
As the method has good results at low time complexity, we decided to not adopt more sophisticated methods. 
Some extracted samples are shown in Table~\ref{tab:sample_acronym_pair} in the appendix.
A manual evaluation on a random sample of 100 extracted sentences yields a precision of 94\%.

\begin{table}[ht] 
	\centering 
	\small
	\setlength{\tabcolsep}{2.2mm}{
		\begin{threeparttable} 
			\begin{tabular}{cccc}  
				\toprule  
				&Short Form&Long Form&Avg\cr
				\midrule
				SciAD \shortcite{veyseh2020does} &732&2,308&3.15\cr
				MadDog \shortcite{veyseh2021maddog}&426,389&3,781,739&8.87\cr
				Ours&1,542,819&6,381,257&4.14\cr
				\bottomrule
			\end{tabular}
			\caption{Statistics for three acronym dictionaries. The ``Avg'' column shows the average number of long forms per acronym.}%
			\label{tab:stat_dict}
		\end{threeparttable}
	}
\end{table}

\textbf{Dictionary Construction.} 
Next, we build a large-scale acronym dictionary with frequencies (popularity) by merging the extracted outputs. 
This merger may regroup duplicate long forms for an acronym, e.g., \textit{``convolutional neural network''}, \textit{``convolutional-neural network''} or \textit{``convolutional neural networks''}.
Therefore, we merge long forms that are identical after stemming and removing punctuation. In our case, the above three forms are merged into \textit{``convolutional neural network''}.
We keep the most frequent, unpreprocessed, long form as the canonical name in our dictionary, discarding other forms.
There are still some noisy long forms that cannot be merged, caused by typos and nested acronyms (see Section~\ref{sec:limitations}). However, a manual evaluation on a sample shows that 94\% of the long forms are clean. If the long forms are weighted by their frequency, the percentage of clean forms increases to 97\%. Most notably, all most frequent long forms for a given acronym were clean in our sample.
The statistics of our dataset are shown in Table~\ref{tab:stat_dict}. Our resource will be the largest publicly available dictionary for acronyms that covers various domains.

\textbf{Pre-training Corpus.} 
While building the dictionary, we can also collect the sentences that contain acronyms for pre-training.
For example, the following sentence contains the acronym \textit{ELEC}: \textit{``Christie, some legislators and the state Election Law Enforcement Commission (\textbf{ELEC}), have joined the comptroller in voicing support for the elimination of the loophole.''} 
For pre-training, the long form \textit{Election Law Enforcement Commission} is removed, and we then force the model to restore the long form from our constructed dictionary, based on the input sentence and the acronym.
In total, we collect a pre-training corpus with \textasciitilde160 million sentences.
More examples are shown in Table~\ref{tab:sample_acronym_pair}.

\begin{figure*}[t]
	\centering
	\includegraphics[width=1.0\textwidth]{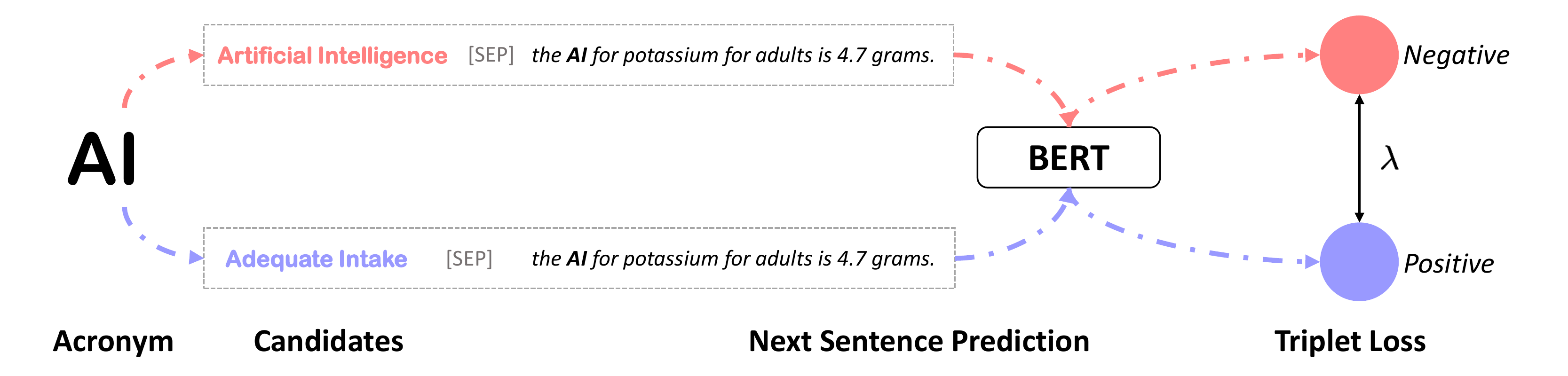}
	\caption{The pre-training strategy of AcroBERT. $\lambda$ is a margin between positive and negative pairs, here
	\emph{$\langle$Adequate Intake, AI$\rangle$}
	and \emph{$\langle$Artificial Intelligence, AI$\rangle$}.}
	\label{fig:framework}
\end{figure*}

\subsection{Acronym Disambiguation Dataset}\label{sec:benchmark}
We use our acronym dictionary to construct new, larger datasets for evaluating AD systems. 
To automatically construct the datasets, we adapt the existing two Entity Disambiguation datasets by replacing the long form of entity with the acronym.
For example, one sentence in Medmentions~\cite{mohan2018medmentions} contains the long form of Cerebral Blood Flow:
\textit{``The reconstructed volume was then compared with corresponding magnetic resonance images demonstrating that the volume of reduced \textbf{Cerebral Blood Flow} agrees with the infarct zone at twenty-four hours''}. The dataset provides the unique ID of this long form in UMLS (\textit{C1623258}), and we use it to find the acronym \emph{CBF} in UMLS.
Therefore, a new sample can be obtained by replacing the long form with its corresponding acronym.

Specifically, we use the following human-annotated and crowd-sourced datasets:\\
\textbf{WikilinksNED Unseen Mentions} \cite{onoe2020fine} is an Entity Disambiguation dataset, i.e., a set of text documents that have mentions of entities, together with a reference knowledge base (KB) that contains, for each entity, one or several names. WikilinksNED Unseen Mentions re-splits the WikilinksNED dataset \cite{eshel2017named} to ensure that all mentions in the validation and test sets do not appear in the training set.
This is a large-scale, crowd-sourced ED dataset from websites in various fields, which is significantly noisier and more challenging than prior datasets. The reference KB is Wikidata (or Wikipedia), and we adapt this WikilinksNED Unseen Mentions to an AD dataset in the \textbf{general} domain.  \\
\textbf{Medmentions}  \cite{mohan2018medmentions} is an entity disambiguation dataset of 4,392 PubMed papers that were annotated by professional and experienced annotators in the biomedical domain. The reference KB is UMLS \cite{bodenreider2004unified}, and this is a \textbf{biomedical} dataset.\\
\textbf{SciAD} \cite{veyseh2020does} is the previously mentioned acronym disambiguation dataset in the \textbf{scientific} domain.

SciAD is already an AD dataset, and we only re-split it to avoid data leakage.
As for the two ED datasets, they both provide a unique ID to the reference KB for each long form.
We then replace the long forms with the acronyms from their corresponding reference KB, i.e., Wikidata and UMLS.
To make sure this replacement is correct, we apply the rule-based algorithm~\cite{schwartz2002simple} to the pair of long form and acronym again for verification.
We manually checked 100 random sentences constructed in this way and did not find problematic cases.
Hence, this semi-synthetic construction results in a dataset of natural text in which the long form and the acronym are mutually replaceable in the context.
\ignore{
\emph{\textbf{the long form and its acronym are mutually replaceable in the context, and therefore the two ED dataset constructions can be automated without introducing noise.}}\gv{I understand that you are using bold because you are trying to address a concern that our previous reviewers had. However, I think that this is counter productive: it will just draw attention to a potential concern.}}
\gv{Suggestion of rephrasing: "This semi-synthetic dataset construction results in a dataset of natural text in which the long form and the acronym are mutually replaceable in the context"} 
\lihu{make sense, I adopted your sentence}
Besides, the pair is added to our dictionary with a frequency of 1 if it does not appear in our dictionary. 
For the WikilinksNED dataset, we use the taxonomy of YAGO~4 \cite{pellissier2020yago} to label each long form with a top-level class. For example, \textit{``rhythm and blues''} is a \texttt{CreativeWork} and \textit{``United States Navy''} is an \texttt{Organization}. 

We then partition the three datasets separately into training, test, and validation set, ensuring that the acronyms in the training set do not appear in the validation and test sets.  
We repartition the datasets at the ratio of 6:2:2.
Table~\ref{tab:stat_benchmark} gives the statistics of this new benchmark. It is not only larger but also more challenging than existing benchmarks, because acronyms in our benchmark have more than 200 candidates on average.
Moreover, it contains many \emph{overshadowed forms}~\cite{provatorova2021robustness}, which means that an acronym has to be disambiguated to a long form that is not the most popular long form for that acronym.
For example, \textit{``Adequate Intake''} is overshadowed by the more popular form \textit{``Artificial Intelligence''} for the acronym \emph{``AI''}.

\begin{table*}[hbt] 
	\centering
	\scriptsize
	\begin{threeparttable} 
		\begin{tabular}{c|cccc|cccc|cccc|ccc}  
			\toprule  Model&\multicolumn{4}{c|}{\textbf{General}}&\multicolumn{4}{c|}{\textbf{Scientific}}&\multicolumn{4}{c|}{\textbf{Biomedical}}&\multicolumn{2}{c}{\textbf{Avg}}\cr
			&\multicolumn{2}{c}{Dev}&\multicolumn{2}{c|}{Test}&\multicolumn{2}{c}{Dev}&\multicolumn{2}{c|}{Test}&\multicolumn{2}{c}{Dev}&\multicolumn{2}{c|}{Test}&\cr
			&F1&Acc&F1&Acc&F1&Acc&F1&Acc&F1&Acc&F1&Acc&F1&Acc\cr
			\midrule
			BM25 \shortcite{robertson1995okapi}&29.9 &32.6&  35.5 &25.8 & 14.1 &5.4 &  17.1 & 10.7&  13.1 & 8.3&  17.0 &14.3 & 21.1 &16.2\cr
			FastText \shortcite{bojanowski2017enriching} &11.3& 12.9& 18.7&12.7 &  3.3&0.9 & 5.7&2.5 & 0.2&0.1 & 1.3&0.7 &6.8&5.0\cr
			MadDog \shortcite{veyseh2021maddog} &28.1&11.7 & 29.9 &  23.1 &  17.8 &  15.5 &  22.4&17.9&33.8&19.3&41.2&35.9&28.9&20.6\cr
			BERT \shortcite{devlin2018bert} & 32.3&32.5&  37.7 &28.2& 15.1&5.8 & 17.6&9.3 & 3.1&1.3 &  3.5&2.1 &  18.2&13.2 \cr
			Popularity-Ours  & 35.2&39.1& 39.0&43.2 &  5.5&22.9 &  4.9&12.3 &  46.0&61.3 &  49.9&54.0 &  30.1&38.8\cr
			AcroBERT  &\textbf{74.7}& \textbf{78.8}&\textbf{70.0}&\textbf{72.0}  & \textbf{26.9} &\textbf{36.6} & \textbf{28.8} &\textbf{27.4}&  \textbf{58.4}&\textbf{66.0}& \textbf{59.9}&\textbf{61.4} & \textbf{53.1}&\textbf{57.0} \cr
			\bottomrule  
		\end{tabular}
		\caption{Performances of the unsupervised setting across different models, measured by macro F1 and Accuracy. }	
		\label{tab:unsupervised_results}
	\end{threeparttable} 
\end{table*}

\section{AcroBERT}\label{sec:acrobert}

We can now capitalize on our dictionary and pre-training corpus to propose a new method for acronym disambiguation. It takes as input (1) a dictionary of acronyms with their long form(s), and (2) a large-scale corpus that contains acronyms (we assume that the boundaries of the acronym have already been recognized). Our goal is to pre-train a language model for acronym disambiguation, which has a strong generalizability across multiple domains.

\ignore{
Formally speaking, the input of this task is a sequence of $n$ words $\mathbf{S} = \langle w_{1}, w_{2}, ..., w_{n} \rangle $, where $w^{acronym}$ is an acronym in the sentence.
In addition, there is an acronym dictionary $\mathcal{D}: A \rightarrow 2^L$, where $A$ is the set of acronyms (with $w^{acronym} \in A$), and $L$ is the set of acronym long forms. The output of our model is an acronym-definition pair $\langle{}w^{acronym}, d\rangle$ with $d \in \mathcal{D}(w^{acronym})$.\fms{Maybe this should all come earlier, in a preliminaries section? Or maybe we don't need one?}
}

\paragraph{The Pre-training Strategy of BERT.} 
We adapt the BERT model for our purpose.
BERT is pre-trained by  using two unsupervised tasks, Masked
Language Model (MLM) and Next Sentence Prediction (NSP).
The Masked Language Model task randomly masks some percentage of the input
tokens, and then forces the model to predict the masked tokens, similar to a cloze task. 
The Next Sentence Prediction task asks the model to predict whether one sentence follows the other.

The Next Sentence Prediction task can be used to predict, from the input text (e.g., \textit{``This is the product's first true \textbf{AI} version, and it understands your voice instantly.''}), the correct long form (\textit{``Artificial Intelligence''}).
Here, 
the model learns to judge whether the input context that contains the acronym \textit{``AI''} is coherent with the long form \textit{``Artificial Intelligence''}. 
The Masked
Language Model task can memorize the correlation of tokens between the context sequence and long form. Thus, the model learns 
that the phrase \textit{``Artificial Intelligence''}  often co-occurs with \textit{``product''} or \textit{``understand''}.

However, we find that this naive technique does not perform well (see the ablation studies in Table~\ref{tab:ablation}).
We believe that the reason is that the acronym is usually ambiguous with many candidates (as shown in Table~\ref{tab:example_of_ai}), so that the model has difficulties predicting the correct long form by only using the cross-entropy loss of the binary classification.
We also observe that the Masked
Language Model loss is so small that the model focuses on adapting the Next Sentence Prediction task only.

\paragraph{AcroBERT.} To mitigate the weaknesses of the original BERT, we pre-train an adapted BERT, called AcroBERT, by slightly adjusting the Next Sentence Prediction task. The framework is shown in Figure~\ref{fig:framework}. It aims to bring the positive sample pairs closer together, and to push apart the negative sample pairs.
We find that already such a simple model can perform very well. 
For each pair of a candidate long form and a sentence with an acronym, we compose an input for the Next Sentence Prediction task as \textit{``\texttt{[CLS]} long form \texttt{[SEP]} sentence \texttt{[SEP]}''}. 
Then, we obtain representations of this sequence by applying BERT to this new input. The final hidden vector  $\textbf{e}^\texttt{[CLS]} \in \mathbb{R}^{H}$ of the first input token (\texttt{[CLS]}) is used as the aggregate representation, where $H$ is the dimension of the hidden vector. 
Next, the scores for the binary classification are:
\begin{equation}
\text{P}(y) = \text{softmax}(\textbf{e}^\texttt{[CLS]} \mathbf{W}), y \in \{0, 1\}
\end{equation}
where $\mathbf{W} \in \mathbb{R}^{H \times 2}$ is a trainable matrix initialized with the weights of the original BERT, and the label $0$ signifies that this pair of sentences are coherent. 
We use $d = \text{P}(y=1)$ as the distance between the candidate and the context,
and we want the distances of negative pairs to be larger than for positive pairs.
For this, we use a triplet loss function that aims to assign higher scores to the correct candidates that match the topic of the input sentence while reducing the scores of irrelevant candidates: 
\begin{equation}
\mathcal{L} = \max \left \{ 0, \lambda - d_{\text{neg} } + d_{\text{pos} } \right \} 
\end{equation}
where $\lambda$ is the margin value, and $d_{\text{pos}}$ and $d_{\text{neg}}$ are the distances for positive and negative pairs, respectively. 

The negatives in this triplet framework can be randomly sampled from the dictionary. 
However, we observe that such random negatives contribute less to the training and result in slower convergence because the initial model can easily distinguish these triplets.
Therefore, it is crucial to select harder triplets that are active and beneficial to the training. For this purpose, we introduce a certain number of ambiguous negatives to each mini-batch,  e.g., \textit{``Artificial Intelligence''} can be added to the input context as an ambiguous negative sample for the positive pair \textit{``Adequate Intake \texttt{[SEP]} In the United States, the \textbf{AI} for potassium for adults is 4.7 grams.''}
Through the pre-training step, AcroBERT is able to identify the correct long form with the most consistent theme from numerous candidates based on the input context.

\begin{table*}[hbt] 
	\centering 
	\scriptsize
	\begin{threeparttable} 
		\begin{tabular}{c|cccc|cccc|cccc|cc}  
			\toprule  &\multicolumn{4}{c|}{\textbf{General}}&\multicolumn{4}{c|}{\textbf{Scientific}}&\multicolumn{4}{c|}{\textbf{Biomedical}}&\multicolumn{2}{c}{\textbf{Avg}}\cr
			&\multicolumn{2}{c}{Dev}&\multicolumn{2}{c|}{Test}&\multicolumn{2}{c}{Dev}&\multicolumn{2}{c|}{Test}&\multicolumn{2}{c}{Dev}&\multicolumn{2}{c|}{Test}&\cr
			&F1&Acc&F1&Acc&F1&Acc&F1&Acc&F1&Acc&F1&Acc&F1&Acc\cr
			\midrule
			BERT \shortcite{devlin2018bert} & 53.8&70.7& 54.9&53.1 & 13.5 &9.9& 14.3 &10.4& 9.8&12.4 &  9.4 &7.5&  26.0&27.3 \cr
			SciBERT \shortcite{beltagy2019scibert}  &32.4&38.6& 33.6 &27.7& 22.7 & 19.4 & 23.4&17.7& 31.2 &35.6&  31.0&28.3 & 29.5&27.9\cr
			BioBERT \shortcite{lee2020biobert}  & 26.0&23.6& 25.7&20.3 &  11.2 &9.7&12.4&9.0&  24.0&21.8 &  20.2&16.8 & 19.9&16.9\cr
			AcroBERT  &\textbf{72.9}&\textbf{76.1}&\textbf{71.0} &\textbf{76.1} & \textbf{28.7} &\textbf{34.9}& \textbf{29.0} &\textbf{27.6}&  \textbf{62.5}&\textbf{62.4} & \textbf{60.3} &\textbf{69.2}& \textbf{54.1}&\textbf{57.7} \cr
			\bottomrule  
		\end{tabular}
		\caption{Performances of fine-tuned setting across different models, measured by macro F1 and Accuracy. }	
		\label{tab:fine-tuning_results}
	\end{threeparttable} 
\end{table*}

\section{Experiments}\label{sec:experiment}

In this section, we compare AcroBERT empirically to other acronym disambiguation approaches.

\subsection{Experimental Settings}
\paragraph{Datasets.} We use the following datasets for evaluation: Our GLADIS benchmark consists of three subsets covering the General, Scientific, and Biomedical domains.
This benchmark is more challenging than prior work due to a large number of ambiguous long forms: each acronym has around 200 candidates on average. We also evaluate AcroBERT on two existing datasets: UAD \cite{ciosici2019unsupervised} and SciAd \cite{veyseh2020does}. They are general and scientific AD datasets, respectively. We reuse the test set of Medmentions but use the UMLS as the target dictionary\fms{Which dataset do we introduce? We build it? Or we use someone's dataset?}.
We refer to them as \textbf{datasets with fewer candidates} because they have fewer candidates per acronym.
The average numbers of candidates per acronym are 2.1, 3.1, and 34.2, respectively. 
See Appendix~\ref{sec:detail_dataset} for more details on the datasets.

\paragraph{Benchmark Settings.}\label{sec:benchmark_setting} 
We design two benchmark settings for the unsupervised and supervised scenarios respectively.
In the \emph{unsupervised setting}, each model is evaluated on the test sets without access to train and validation sets. 
In the \emph{fine-tuned setting}, each model is first fine-tuned on train sets and then evaluated on test sets. 
We focus on the unsupervised setting because it demonstrates that AcroBERT can achieve considerable performances across several domains even without any annotated samples.

\paragraph{Competitors.}\label{sec:competitor} 
We compare our approach to the following publicly available competitors: BM25 \cite{robertson1995okapi}, FastText \cite{bojanowski2017enriching}, MadDog \cite{veyseh2021maddog}, BERT \cite{devlin2018bert}, BioBERT \cite{lee2020biobert}, SciBERT \cite{beltagy2019scibert}.
Besides, we introduce a Popularity-Ours baseline that uses the frequency of long forms of our collected pre-training datasets.
We do not compare to general entity linking methods, because prior work has already found that general systems like AIDA~\cite{hoffart2011robust} tend to lag behind acronym disambiguation models by 10-30 absolute percentage points
~\cite{li2018guess}. 
See Appendix~\ref{sec:competitor} for details on the competitors.

\paragraph{Implementation Details.}
All approaches are implemented with PyTorch \cite{paszke2019pytorch} and HuggingFace~\cite{wolf2020transformers} by using one  NVIDIA Tesla V100S PCIe 32 GB Specs.
For pre-training, we use the parameters in the original BERT to initialize AcroBERT, and then pre-trained on our collected datasets with \textasciitilde160 million samples for one epoch. The margin of triplet loss is 0.2 and the number of ambiguous negatives is 2 for each mini-batch.
See more details in Appendix~\ref{sec:implementation}.
\gv{Can we share code, eg as a zip in the upload? In machine-learning conferences, these days, it is considered a good thing.}
\lihu{absolutely! I'm going to upload the code and datasets. After, I'll make everything publicly available}

\paragraph{Inference.}
For the inference stage, every pair of a context sentence and a candidate with the matching short form
in the dictionary constitutes an input to the Next Sentence Prediction task.
The language model produces a score for each candidate and we select the one with the highest score as the final predicted output.

\paragraph{Metrics.}  We evaluate the models by precision, recall, and macro F1. 
These metrics are defined in detail in Section~\ref{sec:metric}

\subsection{Results}

\subsubsection{Overall Performance}

\begin{figure}[tpb]
	\centering
	\includegraphics[width=0.5\textwidth]{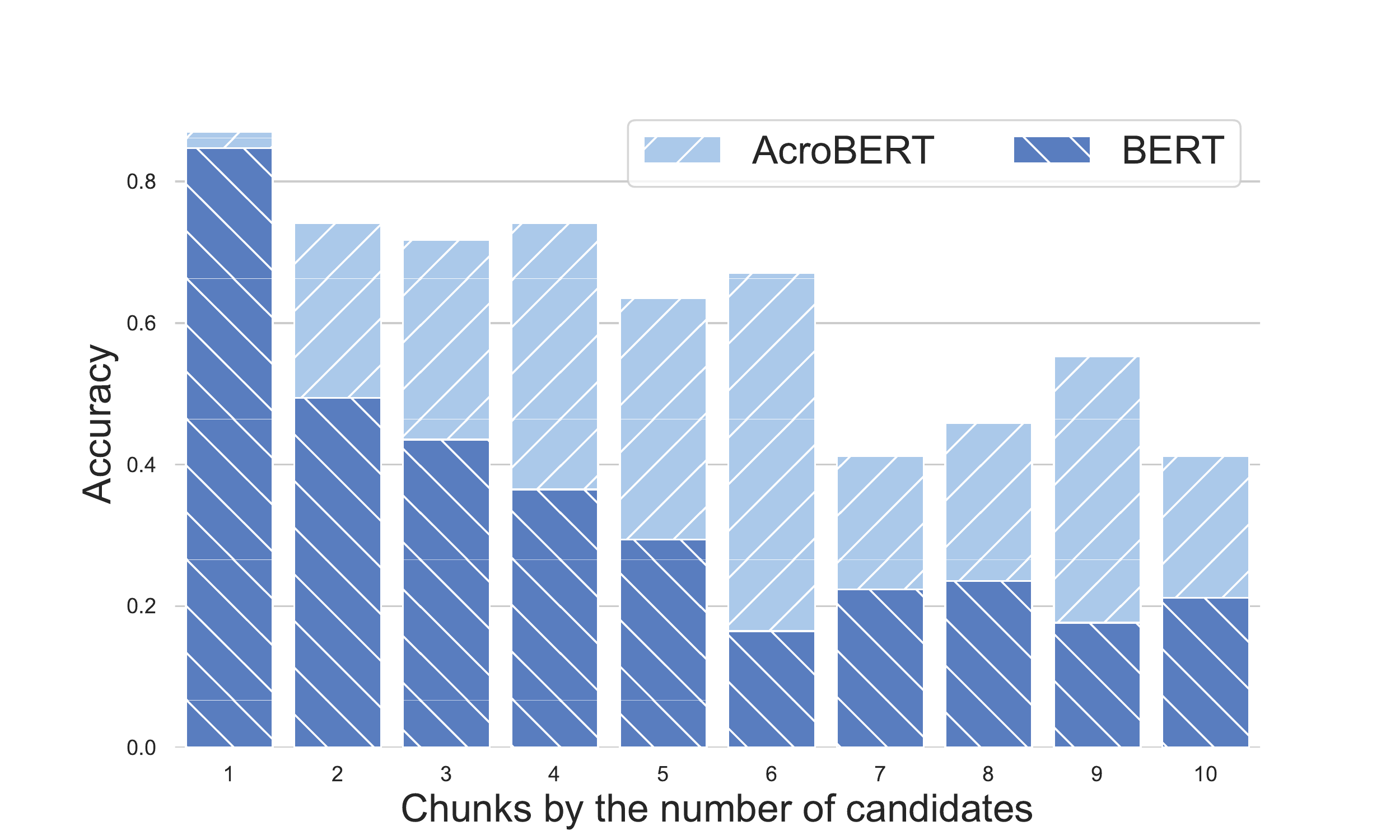}
	\caption{Robustness evaluation of hard samples on the General test set. The samples are divided evenly into ten chunks according to the number of candidates of each sample (results on the other two sets are shown in Figure~\ref{fig:robust_for_two} in the appendix). }
	\label{fig:eval_num_can}
\end{figure}

\textbf{Unsupervised Setting.}
Table~\ref{tab:unsupervised_results} shows the experimental results in the unsupervised setting.
We first observe that our AcroBERT significantly outperforms the baselines across the three domains.
For example, AcroBERT can improve the original BERT by more than 30 absolute percentage points of F1 on average on our benchmark.
Second, the naive popularity method comes second on this benchmark, most likely because it contains a limited number of overshadowed terms. 
However, it performs badly on the scientific dataset. We assume that this is because this dataset contains 68.7\% of overshadowed terms (as shown in Table~\ref{tab:stat_benchmark}).

Besides, we conduct experiments on existing datasets, namely UAD~\cite{ciosici2019unsupervised} and SciAD~\cite{veyseh2020does}. Although our method performs consistently well, we relegate this experiment to the appendix~\ref{sec:fewer_can} due to the weaknesses of the datasets (small size or data leakage).

\textbf{Fine-tuned Setting.}
In this experiment, every pre-trained language model is fine-tuned on the training set by the triplet loss, as introduced in the pre-training step.
Negatives are randomly sampled from ambiguous long forms for the correct label,
and the results are shown in Table~\ref{tab:fine-tuning_results}. 
BERT, SciBERT, and BioBERT perform better in their respective fields.
However, our AcroBERT  achieves the best result across the three fields on average, which demonstrates the effectiveness of the pre-training strategy. 
One might think that it is unfair that AcroBERT uses the pre-training corpus, while the other models do not.
However, there is no other pre-trained model for general disambiguation. Our approach is the first that capitalizes on large-scale corpora and pre-training. 

As for the inference speed, AcroBERT has to be run once for every possible long form, which may take some time if there are thousands of long forms, e.g., the acronym AI. However, this runtime can be reduced drastically if one cuts off the less frequent long forms per acronym. Limiting the number of long forms to 23 per acronym, e.g., reduces the worst-case runtime by a factor of 100, while still keeping the recall at 90\%.

\begin{table}[tbp] 
	\centering 
	\small
	\setlength{\tabcolsep}{2.2mm}{
		\begin{threeparttable} 
			\begin{tabular}{cccc}  
				\toprule  
				\textbf{Model}&\textbf{Popular}&\textbf{Overshadowed}&\textbf{Avg}\cr
				\midrule
				BERT \shortcite{devlin2018bert}&13.3&12.7&13.0\cr
				SciBERT \shortcite{beltagy2019scibert}&11.6&8.1&9.9\cr
				BioBERT \shortcite{lee2020biobert}&2.1&1.0&1.6\cr
				AcroBERT&\textbf{61.9}&\textbf{33.4}&\textbf{47.7}\cr
				\bottomrule
			\end{tabular}
			\caption{Robustness evaluation of overshadowed entities on General test set,measured by Accuracy.} 	\label{tab:overshadowing}
		\end{threeparttable}
	}
\end{table}

\subsubsection{Robustness Evaluation}\label{sec:robust}
Our GLADIS benchmark is more challenging than existing acronym disambiguation datasets due to the much larger acronym dictionary, which means more candidates per acronym. 
To measure the robustness of acronym disambiguation systems against more candidates, we sort the samples in the dataset in descending order of the number of candidates per acronym, and divide them evenly into 10 chunks.
For example, samples in the first chunk have 1.58 candidates on average while that number is 2159 for the last chunk.
The experimental results are shown in Figure~\ref{fig:eval_num_can}.
As expected, the performance of BERT and AcroBERT decreases as the number of candidates increases.
The same goes for the other two subsets, as shown in Appendix~\ref{sec:robust_many_can}. However, AcroBERT consistently outperforms BERT on each data chunk, which shows that AcroBERT is able to select the correct long form among the numerous candidates.

Moreover, the challenge with our GLADIS benchmark comes from overshadowed samples, which are harder to disambiguate.
To validate the robustness of the models, we divide the General test set into two parts: Popular and Overshadowed, as described in Section~\ref{sec:benchmark}.
Next, we compare different language models in the unsupervised setting.
As shown in Table~\ref{tab:overshadowing}, our AcroBERT performs best on both the Popular and the Overshadowed subset.
We conclude that AcroBERT is more robust against ambiguous and overshadowed samples in acronym disambiguation task.

\section{Conclusion}\label{sec:conclusion}
In this paper, we have presented GLADIS, a challenging benchmark for Acronym Disambiguation, which includes a larger dictionary, three datasets from the general, scientific, and biomedical domains, and a large-scale pre-training corpus.
We have also proposed AcroBERT, a BERT-based model that is pre-trained on our collected acronym documents,
which can significantly outperform other baselines across multiple domains, and which is more robust in the presence of very ambiguous acronyms and overshadowed samples.
For future work, we aim to enhance the performance of  AcroBERT on the overshadowed cases, which is crucial for the acronym disambiguation task.

\section{Limitations}\label{sec:limitations}

We see two main limitations of our work. 
First, although the current acronym dictionary is of relatively high quality, it still contains a small fraction of duplicate long forms due to typos (as in \textit{``Convlutional Neural Network''}), morphological changes (as in \textit{``Convolutional Neuronal Network''}) and nested acronyms (as in \textit{``convolutional NN''}). 
A manual evaluation of 100 randomly chosen long forms from the three datasets in GLADIS shows that 6\% of them are noisy.
At the same time, the frequency of these noisy forms is much lower than that of the standard long forms: all noisy forms in the sample taken together appear 100 times in the corpus -- compared to 31k times for the clean forms. Thus, the percentage of clean forms, weighted by their frequency, is 97\%. A good AD system should select the most frequent one among noisy forms for an acronym, and in our sample none of the most frequent forms was noisy.

A second limitation of our approach is that the performance of the current AcroBERT system on the Scientific dataset still needs improvement. We are considering to introduce more pre-training data from this domain to address this issue. 

\section*{Ethics Statement}
This work presents GLADIS, a free and open
benchmark for the research community to study Acronym Disambiguation, which 
consists of three components: a dictionary, a pre-training corpus, and three domain-specific datasets.
The dictionary and pre-training corpus are collected from the Pile dataset~\cite{gao2020pile}, which is a public dataset under the MIT
license.
The three domain-specific datasets are adapted from SciAD~\cite{veyseh2020does}, WikilinksNED Unseen Mentions~\cite{onoe2020fine} and Medmentions~\cite{mohan2018medmentions}, respectively. They all allow sharing
and redistribution. The source datasets and their publications will be credited on our GitHub page, and their licenses will be mentioned both on the Web page and in the downloads of GLADIS.

\section*{Acknowledgements}
This work was partially funded by ANR-20-CHIA-0012-01 (“NoRDF”).
\bibliography{custom}
\bibliographystyle{acl_natbib}

\clearpage
\appendix
\setcounter{table}{0}   
\setcounter{figure}{0}
\renewcommand{\thetable}{A\arabic{table}}
\renewcommand{\thefigure}{A\arabic{figure}}
\setcounter{equation}{0}
\renewcommand{\theequation}{A.\arabic{equation}}
\section{Details of the Experimental Settings}\label{sec:details}
\subsection{Details of the Pile Dataset}\label{sec:pile}
Pile \cite{gao2020pile} is an 825 GiB English text corpus designed to train large-scale language models, which is constructed from 22 diverse high-quality academic or professional sources.
Pile is constructed from existing or newly introduced datasets,
and we present these sources here. 
\textbf{Pile-CC} is a collection of web pages, metadata and texts, which is extracted from jusText \cite{endredy2013more}. 
\textbf{Books3} is a book dataset of fiction and nonfiction books derived from 
Bibliotik \footnote{\url{https://twitter.com/theshawwn/status/1320282149329784833}}.
\textbf{Project Gutenberg} has classic Western literature derived from PG-19 \cite{rae2019compressive}.
\textbf{OpenSubtitles} provides a large corpus of English subtitles from movies and television shows collected by \citet{tiedemann2016finding}.
\textbf{DeepMind Mathematics} is a collection of  many
different types of mathematics questions \cite{saxton2018analysing}.
\textbf{BookCorpus2} is an expanded version of BookCorpus \cite{zhu2015aligning}, a corpus of books from the web.
\textbf{EuroParl} is a corpus of parallel text in 11 languages from the proceedings of the European Parliament\cite{koehn2005europarl}.
\textbf{Enron Emails} is a large set of email messages, which contains 619,446 messages belonging to 158 users \cite{klimt2004enron}.

\begin{figure}[tbp]
	\includegraphics[width=0.5\textwidth]{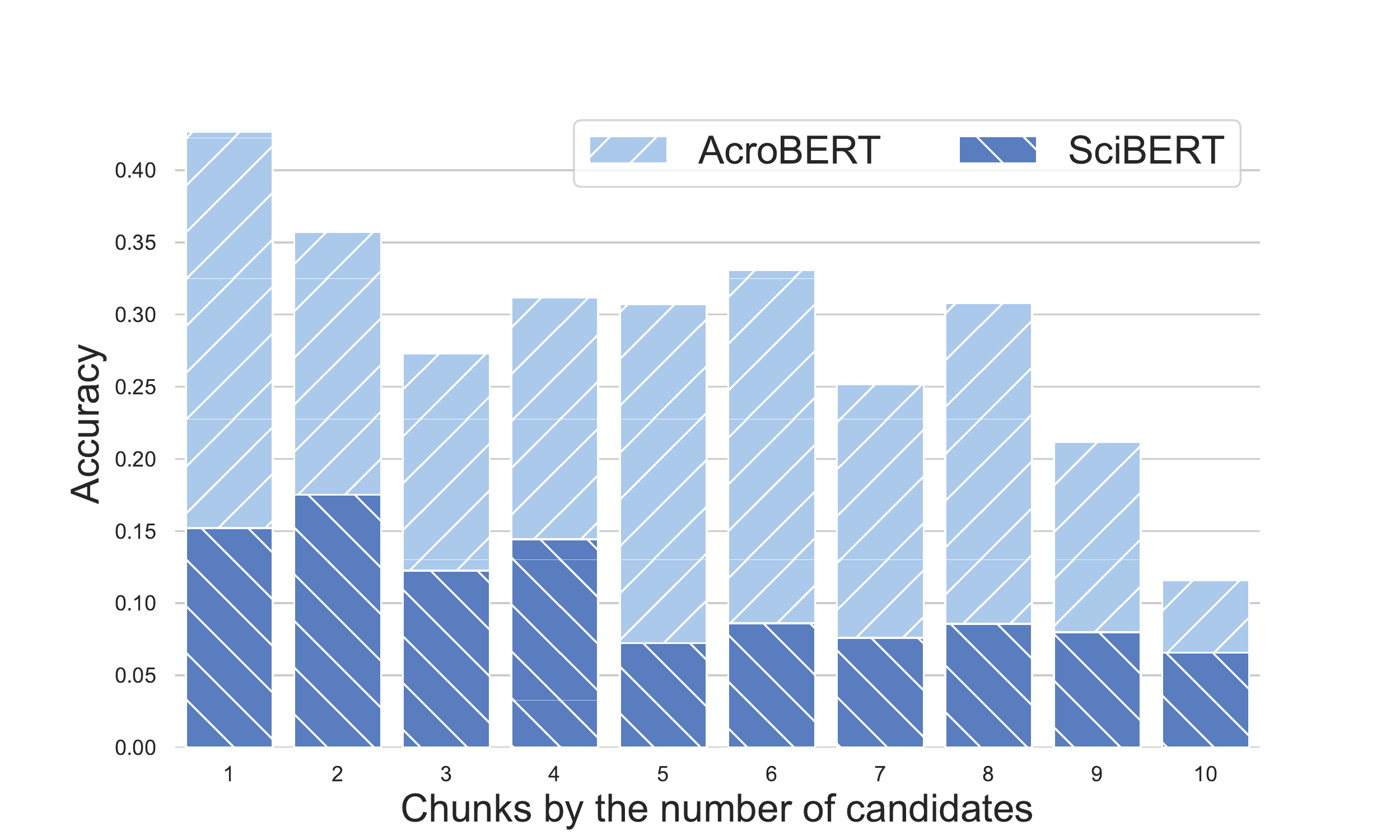}%
	\llap{\raisebox{.27\textwidth}{\sffamily Scientific test set\quad}}
	
	\includegraphics[width=0.5\textwidth]{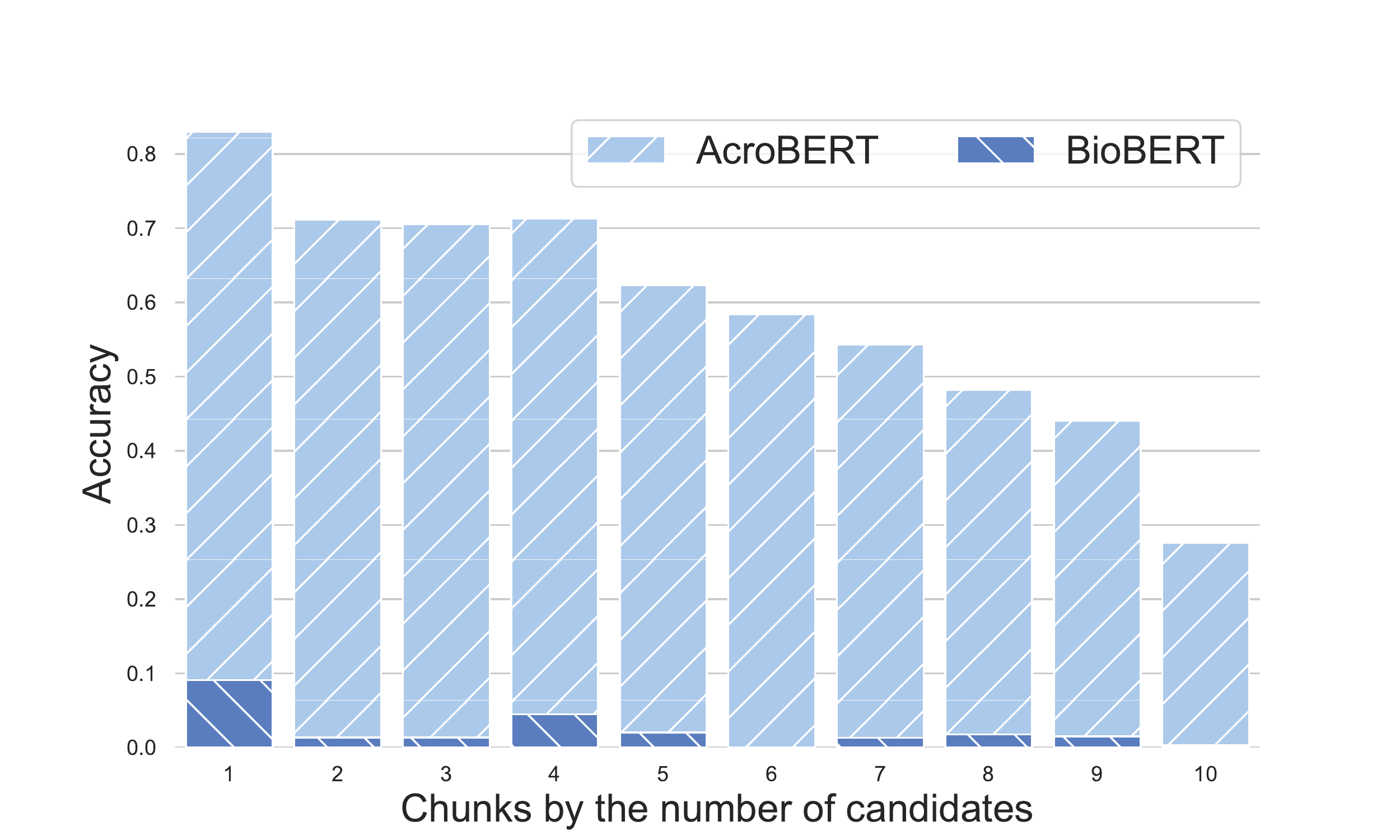}%
	\llap{\raisebox{.27\textwidth}{\sffamily Biomedical test set\quad}}
	\caption{Robustness evaluation of hard samples on Scientific and Biomedical test set. The samples are divided evenly into ten chunks according to the number of candidates of each sample.}
	\label{fig:robust_for_two}
\end{figure}

\subsection{Details of the Experimental Datasets}\label{sec:detail_dataset}
We use the following benchmarks for Acronym Disambiguation:

\begin{table*}[t] 
	\centering\small 
	\setlength{\tabcolsep}{2.2mm}{
		\begin{threeparttable} 
			\begin{tabular}{c|cc|cc|cc|cc}  
				\toprule  
				\textbf{Model}&\multicolumn{2}{c|}{\textbf{UAD}}&\multicolumn{2}{c|}{\textbf{SciAD}}&\multicolumn{2}{c|}{\textbf{Biomedical-UMLS}}&\multicolumn{2}{c}{\textbf{Avg}}\cr
				&F1&Acc&F1&Acc&F1&Acc&F1&Acc\cr
				\midrule
				BERT \shortcite{devlin2018bert} &\textbf{89.3}&91.1&54.1&57.2&38.0&32.7&60.5&60.3\cr
				SciBERT \shortcite{beltagy2019scibert} &74.8&78.4&\textbf{65.6}&71.7&54.9&50.3&65.1&66.8\cr
				BioBERT \shortcite{lee2020biobert} &66.2&68.2&19.7&22.4&37.4&31.4&41.1&40.7\cr
				AcroBERT  &88.8&\textbf{93.7}&58.0&\textbf{72.0}&\textbf{67.5}&\textbf{65.3}&\textbf{71.4}&\textbf{77.0}\cr
				\bottomrule
			\end{tabular}
			\caption{Performances on benchmarks with fewer candidates, measured by macro F1 and Accuracy.} 	\label{tab:other_benchmark}
		\end{threeparttable}
	}
\end{table*}

\textbf{Our GLADIS benchmark} consists of three subsets covering the General, Scientific, and Biomedical domains. It is a very challenging benchmark, due to a large number of ambiguous long forms, as described in Section~\ref{sec:benchmark}. 
\begin{itemize}
	\item \textbf{General} has 45K samples gathered from the WikilinksNED Unseen Mentions \cite{onoe2020fine}.
	\item \textbf{Scientific} is adapted from SciAD~\cite{veyseh2020does} with 56K samples, and the long forms in the original dataset are mapped to the new acronym dictionary. We re-split the training, validation and test sets to assure there are no overlaps.  
	\item \textbf{Biomedical} includes 12K samples obtained from Medmentions \cite{mohan2018medmentions}.
\end{itemize}

\textbf{Datasets with fewer candidates per acronym.}
\begin{itemize}
	\item \textbf{UAD}~\cite{ciosici2019unsupervised} is gathered from the English Wikipedia and we use the manually labeled 7K samples for evaluation.
	\item \textbf{SciAD}~\cite{veyseh2020does} is a human-annotated dataset for the scientific domain with 62K samples gathered from the ArXiv preprint papers, and the validation set with 6K samples is used for experiments.
	\item \textbf{Biomedical-UMLS} is a dataset with 3K samples obtained from the test set in our benchmark by using the UMLS concepts as the acronym dictionary
\end{itemize}
The average candidates per acronym for the three datasets are 2.1, 3.1, and 34.2, respectively.

\subsection{Competitors}\label{sec:competitor} 
We compare our approach to the following publicly available competitors:
\begin{itemize}
	\item BM25 \cite{robertson1995okapi} is a classical ranking function in information retrieval. 
	\item Popularity-Ours is a baseline that uses the frequency of long forms of our collected pre-training datasets. 
	\item BERT \cite{devlin2018bert} is a strong baseline, which pre-trains contextual language models on large corpora.
	The scores for the NSP task can be used for the acronym disambiguation.
	\item FastText \cite{bojanowski2017enriching} is a character-level embeddings and can produce representations for arbitrary words. In this experiment, we first represent the input sentence and candidates by the sum of word embeddings from FastText. Then, all candidates are ranked by their cosine similarity score.   
	\item MadDog \cite{veyseh2021maddog} is a web-based acronym disambiguation system for multiple domains. It first creates chunks in which all samples with the same acronyms are assigned to the same chunks. After, a separate Bi-LSTM model is trained for each chunk. To deploy the MadDog server, it needs at least 125 GB of disk space and 70 GB of RAM memory \footnote{\url{https://github.com/amirveyseh/MadDog}}. 
	\item BioBERT \cite{lee2020biobert} is a biomedical language representation model mainly pre-trained on PubMed Abstracts and PMC Full-text articles, which is a strong baseline in the biomedical domain.
	\item SciBERT \cite{beltagy2019scibert} is a scientific language model based on BERT pre-trained on a large multi-domain corpus of scientific publications, which can improve performance on downstream scientific NLP tasks.
\end{itemize}

\begin{table*}[ht] 
	\centering 
	\small
	\setlength{\tabcolsep}{2.2mm}{
		\begin{threeparttable} 
			\begin{tabular}{l p{3.5cm} p{9.5cm}}  
				\toprule  
				Acronym&Long Form&Provenance\cr
				\midrule
				ELEC&\textit{Election Law Enforcement Commission}&\textit{Christie, some legislators and the state Election Law Enforcement Commission (\textbf{ELEC}), have joined the comptroller in voicing support for the elimination of the loophole.}\cr
				\midrule
				ISR	& \textit{in-stent restenosis}&	\textit{Although conventional stents are routinely used in clinical procedures, clinical data shows that these stents are not capable of completely preventing in-stent restenosis (\textbf{ISR}) or restenosis caused by intimal hyperplasia.}\cr
				\midrule
				IL-6&\textit{interleukin-6}&\textit{Consistent blood markers in afflicted patients are normal to low white cell counts and elevated interleukin-6 (\textbf{IL-6}) levels which, among its many activities, signal the liver to increase synthesis and secretion of CRP.}\cr
				\midrule
				PCP&\textit{Planar cell polarity}&\textit{Establishment of photoreceptor cell polarity and ciliogenesis Planar cell polarity (\textbf{PCP})-associated Prickle genes (Pk1 and Pk2) are tissue polarity genes necessary for the establishment of PCP in Drosophila.}\cr
				\midrule
				DEP&\textit{dielectrophoretic}&\textit{They included: a particle counter, trypan blue exclusion (Cedex), an in situ bulk capacitance probe, an off-line fluorescent flow cytometer, and a prototype dielectrophoretic (\textbf{DEP}) cytometer.}\cr
				\midrule
				AQP3&\textit{aquaporin3}&\textit{The laxative effect of bisacodyl is attributable to decreased aquaporin-3 expression in the colon induced by increased PGE2 secretion from macrophages.The purpose of this study was to investigate the role of aquaporin3 (\textbf{AQP3}) in the colon in the laxative effect of bisacodyl.}\cr
				\bottomrule
			\end{tabular}
			\caption{Samples of extracted acronyms, long forms and provenances by using the rule-based algorithm from \citet{schwartz2002simple}. } 	\label{tab:sample_acronym_pair}
		\end{threeparttable}
	}
\end{table*}

\begin{table*}[ht] 
	\centering 
	\tiny
	\setlength{\tabcolsep}{2.2mm}{
		\begin{threeparttable} 
			\begin{tabular}{c|p{2cm}|p{7cm}|p{2cm}|p{2cm}}  
				\toprule  
				Acronym&Long Form&Context&BERT&AcroBERT\cr
				\midrule
				ECB&\textit{European Central Bank}&\textit{Being made to bring the main road network in Romania in the European corridors. There have been initiated several projects to modernize the network of \textbf{ECB} corridors, financed from ispa funds and state-guaranteed loans from international financial institutions. Government seeks external financing or public-private partnerships for other road network upgrades , especially}&\textcolor{c1}{\textit{External Commercial Borrowing}}&\textit{\textcolor{c2}{European Central Bank}}\cr
				\midrule
				PR&\textit{Public Relations}&\textit{A whistleblower like monologist Mike Daisey gets targeted as a scapegoat who must be discredited and diminished in the public 's eye. More often than not, \textbf{PR} is a preemptive process. Celebrity publicists are paid lots of money to keep certain stories out of the news.}&\textcolor{c1}{\textit{Preemptive-Resume}}&\textit{\textcolor{c2}{Public Relations}}\cr
				\midrule
				PUD&\textit{Peptic Ulcer Disease}&\textit{Tumors cause an overactivation of these hormone-producing glands, leading to serious health problems such as severe \textbf{PUD} ( due to gastrin hypersecretion, which stimulates secretion of hydrochloric acd ). }&\textcolor{c1}{\textit{Psychogenic Urinary Dysfunction}}&\textit{\textcolor{c2}{Peptic Ulcer Disease}}\cr
				\midrule
				WFC&\textit{Walsall F.C.}&\textit{Injury during a game against Norwich city on the 13 march 2010, forcing him to miss Huddersfields next five games. He made his return against \textbf{WFC} on the 13 April 2010 , coming on as a 75th minute substitute and scoring a stoppage time winner to make the score 4a3 to town.  }&\textcolor{c1}{\textit{Wide Free Choice}}&\textit{\textcolor{c1}{World Fighting Championships	}}\cr
				\bottomrule
			\end{tabular}
			\caption{Case study of predicted results by BERT and AcroBERT. }\label{tab:case_study}
		\end{threeparttable}
	}
\end{table*}

\begin{table*}[t] 
	\centering\small 
	\setlength{\tabcolsep}{2.2mm}{
		\begin{threeparttable} 
			\begin{tabular}{ccccccccc}  
				\toprule  
				\textbf{Strategy}&\multicolumn{2}{c}{\textbf{General}}&\multicolumn{2}{c}{\textbf{Scientific}}&\multicolumn{2}{c}{\textbf{Biomedical}}&\multicolumn{2}{c}{\textbf{Avg}}\cr
				&F1&Acc&F1&Acc&F1&Acc&F1&Acc\cr
				\midrule
				Triplet framework in AcroBERT &\textbf{61.2}&\textbf{64.2}&\textbf{18.8}&\textbf{18.4}&9.1&10.1&\textbf{29.7}&\textbf{30.9}\cr
				MLM + NSP of the original BERT   &45.7&55.7&15.8&14.9&\textbf{10.2}&\textbf{11.4}&23.9&27.3\cr
				only NSP &52.4&58.7&15.1&12.4&6.1&6.5&24.5&25.9\cr
				only MLM &11.2&14.4&2.7&3.5&1.4&1.5&5.1&6.5\cr
				\bottomrule
			\end{tabular}
			\caption{Ablation studies for different pre-training strategies after 300K steps across three validation sets, measured by macro F1 and Accuracy.} 	\label{tab:ablation}
		\end{threeparttable}
	}
\end{table*}

\subsection{Implementation Details} \label{sec:implementation}
All approaches are implemented with PyTorch \cite{paszke2019pytorch} and HuggingFace~\cite{wolf2020transformers}. 
When pre-training AcroBERT, the model is initialized by the parameters in the original BERT, and then pre-trained on our collected datasets for one epoch. 
In total, there are \textasciitilde160 million samples in this corpus, covering various domains.
The batch size is 32, and we use Adam \cite{kingma2014adam} with a learning rate $2e-5$ for optimization. The learning rate is exponentially decayed for every 10,000 steps with a rate of 0.95.
The margin of triplet loss is 0.2 and the number of ambiguous negatives is 2 for each mini-batch.

For the fine-tuning stage, each competitor model is initialized with the pre-trained parameters from HuggingFace, and we use AcroBERT after pre-training for comparison.
All models are fine-tuned by using the Triplet loss.
All parameters of each model are fine-tuned in this experiment, across all domains by using the same hyper-parameters. The batch size is 8 and the learning rate is among $[1e-5, 8e-6, 6e-6, 4e-6, 2e-6]$ for the Adam optimizer. The model that has the best performance on the validation set among the 5 learning rates is evaluated on the test set. We use one NVIDIA Tesla V100S PCIe 32 GB Specs.

\subsection{Metrics}\label{sec:metric}
Acronym disambiguation can be seen as a classification problem, where the input is (1) a dictionary of acronyms and (2) a sentence with an acronym. Each long form for that acronym from the dictionary is considered a class, and the acronym disambiguation has to choose the correct class. We evaluate the models by precision, recall, and macro F1. 
There are two ways to calculate the macro F1: ``F1 of Averages'' and ``Averaged F1''.
The first computes the F1 value over the arithmetic means of precision and recall, while the second computes the F1 value for each class, and then averages them. Some prior works adopt the first method. However, this method gives a higher weight to popular classes, and it may thus unfairly yield a high score if the model works well on these popular classes only
\cite{opitz2019macro}.
Therefore, we use the Averaged F1 across classes as our metric, which is more robust towards the error type distribution. That is:
\begin{gather}
	\text{Precision}_{i}= \frac{\text{TP}_{i}}{\text{TP}_{i} + \text{FP}_{i}}, i \in \{1,2,...,n\}\\
	\text{Recall}_{i}= \frac{\text{TP}_{i}}{\text{TP}_{i} + \text{FN}_{i}}, i \in \{1,2,...,n\}\\
	\text{F1}= \frac{1}{n} \sum_{i=1}^{n}\frac{2 \times \text{Precision}_{i} \times \text{Recall}_{i}}{\text{Precision}_{i} + \text{Recall}_{i}}
\end{gather}

\section{Additional Experiments}\label{sec:additional}

\subsection{Ablation Study}
In this experiment, we validate the effectiveness of the pre-training strategy in AcroBERT, which adopts a triplet loss with negative samples from ambiguous candidates.
Every model is initialized with the parameters of the original BERT, and 
we use various strategies for pre-training: only Masked Language Model, only Next Sentence Prediction, and the combination of the two and the triplet framework in AcroBERT.
Each strategy is pre-trained on our collected corpus for 300K steps and then the corresponding model is evaluated on the three validation sets.
The results (in Table~\ref{tab:ablation}) show that the strategy of AcroBERT is most beneficial for the acronym disambiguation, as it performs the best on average. 
The Next Sentence Prediction task is more important than the Masked Language Model task.
Even if MLM is not used, the impact on the model is not significant, which means
the original BERT has already learned it well.

\subsection{Experiments on Benchmarks with Fewer Candidates} \label{sec:fewer_can}
As mentioned before, one drawback of the prior AD benchmark is that the magnitude of the acronym dictionary is small, which is not consistent with practical applications. 
In this experiment, we therefore valid the performance of AcroBERT on datasets with fewer candidates.
The results are shown in Table~\ref{tab:other_benchmark}, and we observe that AcroBERT can achieve the best average scores across three datasets again, which demonstrates the generalization capability of our AcroBERT.
On the other hand, the lead of our model is not as substantial as before. This is because there are fewer candidates per acronym, 
and AcroBERT is particularly well-suited for identifying the correct one among a large number of candidates.

\subsection{Robustness Evaluation for Many Candidates} \label{sec:robust_many_can}
Similar to Section~\ref{sec:robust}, we analyse the robustness of AcroBERT on the other two domains.
Each test set is divided evenly into 10 chunks by the number of candidates. The first chunk has the least number of candidates while the last chunk has the most, up to more than 2K. 
Figure~\ref{fig:robust_for_two} shows the experimental scores on the Scientific and Biomedical test set. 
We observe that for the first chunk, SciBERT and BioBERT are on par with our AcroBERT. However, AcroBERT outperforms the two significantly when the number of candidates get larger. 

\subsection{Case Study}
Table~\ref{tab:case_study} shows case studies of the outputs by BERT and AcroBERT. 
BERT often uses the memorized correlations of tokens for reasoning and this can cause errors.
For example, \textit{External Commercial Borrowings} are loans in India made by non-resident lenders in foreign currency to Indian borrowers \footnote{\url{https://en.wikipedia.org/wiki/External_commercial_borrowing}}. 
BERT can determine this correct long form probably with help of the key phrase \textit{``external financing''}.
For the third case, \textit{Peptic Ulcer Disease} is more consistent with the input context. However, BERT fails on it while AcroBERT benefits from the pre-training strategy and is able to distinguish different candidates based on contexts.
For the fourth sample, both methods fail, most likely because of the low frequency of the long forms and the uninformative contexts.
\end{document}